\documentclass{article} % For LaTeX2e
\usepackage{collas2026_conference,times}
\usepackage{easyReview}

\usepackage{amsmath,amsfonts,bm}

\def\eqref#1{equation~\ref{#1}}
\def\1{\bm{1}}

\DeclareMathAlphabet{\mathsfit}{\encodingdefault}{\sfdefault}{m}{sl}
\SetMathAlphabet{\mathsfit}{bold}{\encodingdefault}{\sfdefault}{bx}{n}

\usepackage{hyperref}
\hypersetup{
    colorlinks=true,
    linkcolor=red,
    filecolor=magenta,
    urlcolor=blue,
    citecolor=purple,
    }

\usepackage{enumitem}
\usepackage{graphicx}
\usepackage{booktabs}
\usepackage{subcaption}

\title{Forgetting, plasticity, \underline{and} co-observation:\\a third facet of continual learning}
\author{Timm Hess
\\
ESAT\,\textsuperscript{$\dagger$}\\
KU Leuven, Belgium \\
\And % Use And to have authors side by side
Abhishek Jha\\
VIB.AI\,\textsuperscript{$\dagger$}\\
KU Leuven, Belgium \\
\And
Gido M.\ van de Ven\\
Bernoulli Institute\,\textsuperscript{$\ddagger$}\\
University of Groningen, Netherlands\\
\And
Tinne Tuytelaars\\
ESAT\,\textsuperscript{$\dagger$}\\
KU Leuven, Belgium \\
}

\usepackage{xcolor}
\definecolor{darkblue}{rgb}{0.0, 0.0, 0.5}
\definecolor{darkblue}{rgb}{0.0, 0.0, 0.5}
\definecolor{darkgreen}{rgb}{0.0, 0.5, 0.0}
\definecolor{darkorange}{rgb}{0.85, 0.4, 0.0}

\usepackage{pifont}
\newcommand{\cmark}{\textcolor{darkgreen}{\ding{51}}}
\newcommand{\xmark}{\textcolor{red}{\ding{55}}}

\collasfinalcopy % Uncomment for camera-ready version, but NOT for submission.

\begin{document}

\maketitle
\let\thefootnote\relax\footnotetext{%
  \textsuperscript{$\dagger$}\texttt{\{timmfelix.hess, abhishek.jha, tinne.tuytelaars\}@kuleuven.be; \textsuperscript{$\ddagger$}g.m.van.de.ven@rug.nl}
}

\begin{abstract}
    Efficient continual learning remains a fundamental challenge for deep neural networks. While catastrophic forgetting and loss of plasticity are widely considered the primary obstacles to overcome, we show that these two issues cannot fully explain the performance gap between naive sequential training and offline joint training. %argue that even perfectly resolving these two issues falls short matching the generalization capabilities of (offline) joint training.
    In this paper, we highlight data \textit{co-observation} as a distinct factor influencing continual learning performance. By decoupling the constraints of \emph{separate} data access from stability and plasticity, we systematically investigate the representational benefits gained by observing training data together. Empirically, we demonstrate a consistent performance difference between joint and separate training across both supervised and self-supervised paradigms in generic data-incremental ``chunking" scenarios, whilst mitigating forgetting and controlling for plasticity. Our findings indicate that simultaneous observation of training data (co-observation) yields benefits to the learner's generalization that extend well beyond mere knowledge retention, and that this effect does not require a specific continual distribution shift. Furthermore, we contextualize prominent continual learning mechanisms through this lens: while distillation-based approaches act only as effective knowledge retention mechanisms, our results suggest that the empirical success of memory replay goes beyond the mitigation of forgetting, actively reintroducing the benefits of data co-observation into the learning process.
\end{abstract}

\section{Introduction}
Deep neural networks are predominantly trained under a joint training paradigm, where the model is optimized on independent and identically distributed (i.i.d.)\ batches drawn uniformly from a static dataset. In many real-world settings, however, it would be transformative if this paradigm need not be maintained. When new data arrives sequentially over time, retraining on all previously observed data becomes expensive or sometimes infeasible due to, e.g.\ computational, privacy, or storage constraints \citep{verwimp2024continual}. The model could instead be sequentially trained on separate partitions of data as the data arrives.
This process of incrementally updating a model from a non-stationary data stream is referred to as continual learning (CL) or lifelong learning~\citep{chen2018lifelong, parisi2019continual}.
A fundamental challenge in this regime stems from what is known as the stability-plasticity dilemma~\citep{grossberg1980does, mermillod2013stability}. On the one hand, an effective continual learning system requires sufficient \textit{plasticity} to adapt its representation based on new data. On the other hand, updating the model exclusively on new observations disrupts existing parameter configurations thereby severely impairing its \textit{stability}, a.k.a.\ catastrophic forgetting~\citep{mccloskey1989catastrophic, parisi2019continual}. Consequently, a large part of the continual learning literature focuses on algorithmic interventions designed to retain prior knowledge while remaining capable of incorporating new information~\citep{de2021continual, wang2024comprehensive}.

Progress on these two desiderata is typically assessed relative to a pair of canonical reference points: a lower target, established by naive sequential training, typically interpreted as exposing the severity of forgetting in the absence of any intervention; and an upper target, established by (offline) joint training, interpreted as reflecting the maximum achievable performance on the full dataset. This evaluation framework forwards the implicit assumption that perfectly managing stability and plasticity during sequential training would close the gap to joint training.
In our work, we question this assumption. 
We hypothesize that, under a sequential training regime, even a model with perfect stability and plasticity would still underperform a jointly trained model in terms of its generalization properties. 
This intuition can be linked to multi-task learning theory \citep{caruana1997multitask}, which posits that simultaneous training provides a mutually beneficial inductive bias. 
Our work empirically investigates this dynamic in continual learning, and we demonstrate a benefit of simultaneous training even in identically distributed, data-incremental setups.
To put this another way, we argue that the challenge of sequential training can be split into two parts: the temporal aspect, which underlies forgetting and plasticity, and the challenge of \emph{separate training}, whereby cross-partition dependencies remain hidden. 
We refer to the influence of this phenomenon as the \emph{co-observation effect}.
An analogy, illustrated in Figure~\ref{fig:intuitive_illustration}, provides a simple intuition for the challenge of separate training: assembling a puzzle is easier when all pieces are visible simultaneously, because overarching patterns are more readily recognized when the pieces that jointly encode them are co-observed.

To empirically support our hypothesis, we report our results on data-incremental ``chunking" scenarios~\citep{lee2023chunking}, which avoid macroscopic distribution shifts, arguably making it one of the hardest settings to show this effect. Furthermore, we show that this co-observation effect persists across both supervised and self-supervised learning paradigms. 
We design a diagnostic methodology based on continual pre-training~\citep{hu2022well, zhou2025revisiting} and model ensembles~\citep{vogelstein2020representation} to isolate the effect of co-observation as much as possible.
Inspired by \citet{hess2023knowledge}, we evaluate continual learning at representation level via linear probing~\citep{alain2016understanding}, which decouples the model's generalization from the confounding dynamics of a continually updated classifier head. 
Moreover, we employ an ensemble to control for forgetting. The ensemble preserves the full representational state of the model after each sequential training phase, thereby avoiding forgetting by design. Our results show that in the ``chunking" setting, without distribution shift and class imbalance as additional confounds, the ensemble-baseline is consistently surpassed by incremental joint training. %upper target. 
Because the impact of forgetting is controlled for, we attribute this residual difference in results to the absence of co-observation during optimization.

Ultimately, our findings point towards an important realization for the conceptualization of continual learning mechanisms. Maintaining stability and plasticity is the fundamental requirement in continual learning. To approach the generalization capabilities of joint training, continual learning methods must also acknowledge and compensate for the representational deficit caused by the loss of co-observation in separate training. We summarize our contributions as follows:

\begin{figure}[tb]
    \centering
    \includegraphics[width=0.8\linewidth]{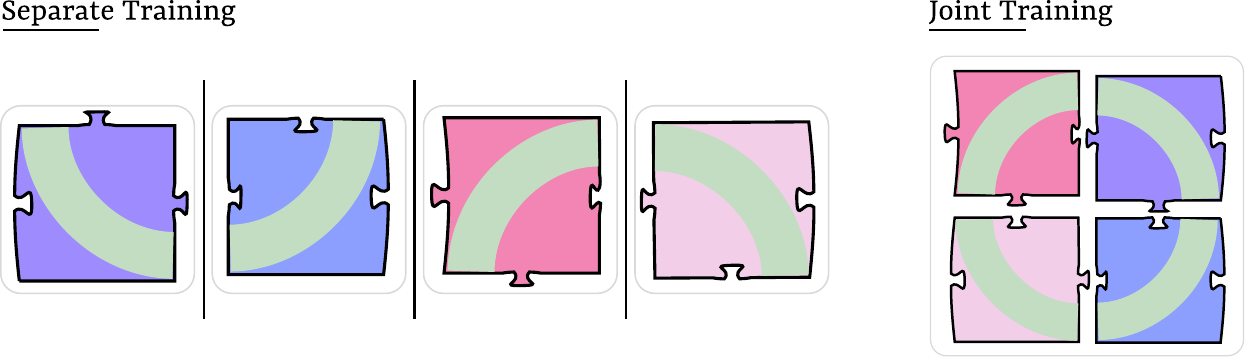}
    \caption{Intuitive illustration of the generalization properties of separate training in comparison to joint training. Finding a common structure across the presented information, here depicted as puzzle pieces, is easier when observing the information together, as in the joint training case (right). In contrast, we can think of separate training as placing information piece-by-piece while focusing on one piece of information at a time (left). Here, local fits emerge that struggle to map relationships across multiple pieces of information.}
    \label{fig:intuitive_illustration}
\end{figure}

\begin{itemize}
    \item \textbf{Conceptualization of the co-observation effect:} We distinguish the impact of separate training from stability-plasticity, demonstrating co-observation as a distinct mechanism when it comes to explaining the gap between continual learning and joint learning.
    \item \textbf{Diagnostic methodology for separating forgetting from co-observation:}\footnote{Code: \url{https://github.com/TimmHess/Co-ob}.} We design an
    experimental setup based on continual pre-training, linear probing, and a full-retention ensemble baseline. This setup effectively disentangles the performance degradation caused by catastrophic forgetting from the representational deficit caused by the missing co-observation of the data during training.
    \item \textbf{Empirical evidence in the vision domain:} We provide rigorous empirical evidence of the co-observation effect across both supervised and self-supervised continual pre-training paradigms, on standard CL vision benchmarks in the ``chunking'' setting, demonstrating that it is a fundamental limitation of separate training.
    \item \textbf{Contextualization of memory replay and knowledge distillation:} We discuss the broader implications of our findings, marking that the empirical success of memory rehearsal methods stems not only from preventing forgetting but also from actively restoring the conditions necessary for data co-observation. And vice versa, knowledge distillation at a representation level alone can compensate forgetting, but not co-observation.
\end{itemize}

\section{Isolating the co-observation effect}
\label{sec:isolating_coob}

To understand the impact of data co-observation in continual learning, it must be carefully detached from the impact of forgetting and loss of plasticity. In this section, we first make a conceptual argument for the co-observation effect, detailing how the lack of simultaneous data access implies a fundamental representational deficit. Subsequently, we introduce our methodology to demonstrate this phenomenon empirically. Because standard end-to-end evaluations conflate the effect from co-observation with effects from stability-plasticity, and come with additional challenges regarding confounding variables such as task-recency bias and class imbalance \citep{wu2019large, davari2022probing}, we design a diagnostic evaluation framework focused on continual representation learning to decouple these effects as cleanly as possible.

\subsection{A conceptual explanation of the co-observation effect}
\label{subsec:conceptual_explanation}
Assume a global data distribution is partitioned into sequential observations, 
$\mathcal{D}_1$ and $\mathcal{D}_2$. And, assume there exists a set of generalizable features, $Z_{\text{shared}}$, which improve prediction of the target variable of interest across the joint distribution. Following multi-task learning theory \citep{baxter2000model}, discovering $Z_{\text{shared}}$ is facilitated by minimizing risk across multiple data variations simultaneously. When optimizing solely on $\mathcal{D}_1$, neural networks naturally favor simpler functions (simplicity bias)~\citep{shah2020pitfalls}. 
Consequently, the model relies on task-specific shortcuts $Z_{\text{local}}$ instead of learning $Z_{\text{shared}}$. 
Because $Z_{\text{shared}}$ was never necessary to minimize the loss in $\mathcal{D}_{1}$, it becomes unidentifiable from the frozen parameters $\theta_{1}$. 
Consider an ideal continual learner transitioning to $\mathcal{D}_2$ with perfect stability (zero forgetting) and plasticity. While it perfectly retains $\theta_1$, these parameters do not encode the discarded $Z_{\text{shared}}$, and when optimizing exclusively on $\mathcal{D}_2$, the model will again favor local shortcuts over discovering $Z_{\text{shared}}$.
We argue that discovering the overarching synergies of $Z_{\text{shared}}$ fundamentally requires updating features based on their simultaneous performance across both distributions. Without concurrent access to both $\mathcal{D}_1$ and $\mathcal{D}_2$, the model lacks the mutual constraints necessary to favor shared features. Thus, even a model that perfectly retains all past parameter states cannot necessarily synthesize cross-distribution dependencies.

\subsection{Continual learning and linear probing}
We consider a standard separate training setting where a model is trained sequentially on a series of data subsets $\mathcal{D}_1, \mathcal{D}_2, \dots, \mathcal{D}_T$. At each training stage $t$, by default, the model has access only to $\mathcal{D}_t$ and cannot observe data from previous or future stages. 
Access to previous data is permitted only when a CL mechanism maintains a memory buffer.

To evaluate the fundamental quality of the knowledge acquired during this process, we decouple the model into a feature extractor (backbone) and a classifier (head), focusing our analysis entirely on the feature extractor. Following standard practices in representation learning, we evaluate all approaches using linear probing~(LP; \citealp{alain2016understanding}). After completing a stage in the sequential training process, we freeze the learned feature extractor and train a linear classifier on top of the frozen representations using all available training data $\bigcup_{t=1}^T \mathcal{D}_t$. 
By evaluating the representation globally, we ensure that any differences in performance reflect its generalizability, rather than artifacts of a continually shifting decision boundary. 

\subsection{Four baselines}
\label{subsec:three_baselines}
\begin{figure}[htb]
    \newlength{\figHTw}
    \setlength{\figHTw}{4.5cm}
    \centering
    \begin{subfigure}[t]{0.36\textwidth}
        \centering
        \includegraphics[height=\figHTw]{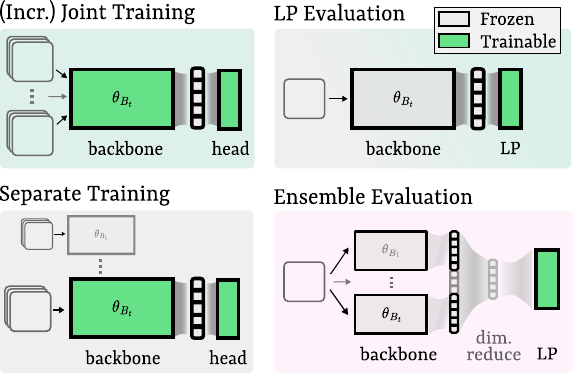}
        \caption{}
    \end{subfigure}
    \hfill
    \begin{subfigure}[t]{0.33\textwidth}
        \centering
        \includegraphics[height=\figHTw]{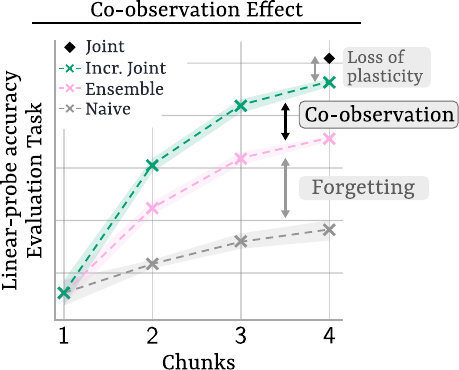}
        \caption{}
    \end{subfigure}
    \begin{subfigure}[t]{0.2\textwidth}
        \centering
        \vspace{-4cm}
        \setlength{\tabcolsep}{4pt}
        \scalebox{0.9}{%
        \begin{tabular}{lccc}
        & \rotatebox{90}{Forgetting} & \rotatebox{90}{Co-observation}  & \rotatebox{90}{Plasticity}\\
        \midrule
        Joint              & \cmark & \cmark & \cmark \\
        Incr. Joint  & \cmark & \cmark & \xmark \\
        Ensemble           & \cmark & \xmark & \xmark \\
        Naive              & \xmark & \xmark & \xmark \\
        \bottomrule
        \end{tabular}
        }
        \vspace{0.35cm}
        \caption{}
        \label{fig:ensemble_explain:right}
    \end{subfigure}
    \caption{\textbf{Overview of training and evaluation for joint and separate training and expected empirical decomposition of the co-observation effect.} \textbf{(a)} In the (incremental) joint training settings (top), the backbone $\theta_{B_t}$ is trained with simultaneous access to all data (accumulated up to stage $t$). The \textit{incremental joint} model is continually updated from the previous stage's initialization at each step, whereas the offline \textit{joint} model is trained from scratch and depicted only at the final stage. In the separate training setting (bottom), the model observes only the current data chunk. To demonstrate the effect of co-observation, we eliminate forgetting without breaking the sequential visibility constraint using the \textit{ensemble} baseline. The ensemble stores a frozen copy of the backbone after each task. During evaluation, representations from all $t$ checkpoints are concatenated, optionally projected back to the original dimension $d$ to control for representational capacity (dim. reduce), and evaluated via linear probing (LP). \textbf{(b)} Hypothesis of the expected results %for training and evaluation 
    across four chunks. The performance difference between the naive sequential model and the ensemble quantifies \textit{forgetting}, while the gap between the incremental joint model and the (offline) joint target highlights the \textit{loss of plasticity}. The focus of this work lies on the remaining difference between the ensemble and the incremental joint model, which we attribute to the \textit{co-observation} effect. \textbf{(c)} Summary of how the four baselines inherently account for the presence (\cmark) or absence (\xmark) of forgetting mitigation, data co-observation, and maintained plasticity.}
    \label{fig:ensemble_explain}
    \vspace*{3em}
\end{figure}

To isolate the influence of co-observation, we compare the sequentially trained representation against an upper target while controlling for catastrophic forgetting. We achieve this by comparing four distinct baselines (\textit{c.f.}~Figure~\ref{fig:ensemble_explain}):
\begin{itemize}
    \item \textbf{Joint (offline):} A single model is trained with simultaneous access to all data $\bigcup_{i=1}^T \mathcal{D}_i$. This represents the standard offline machine learning paradigm, i.e. the common choice of upper target for representation quality.
    \item \textbf{Incremental joint:} At each stage $t$, the model is trained with simultaneous access to all data encountered so far, i.e. $\bigcup_{i=1}^t \mathcal{D}_i$. Because the model trains on all available past and present data simultaneously, it circumvents both fragmented data access and catastrophic forgetting.
    Because it is subject to loss of plasticity from repeated optimization, as other continual learning approaches, we utilize it for fair comparison. 
    \item \textbf{Ensemble (perfect forgetting mitigation):} 
    To eliminate forgetting of previously learned knowledge without explicitly breaking the separate observability constraint of sequential training, we employ an ensemble approach. After each sequential training stage $t$, a copy of the feature extractor is stored. At evaluation time, the representations from all $T$ stored checkpoints are concatenated, and a single linear probe is trained on this combined representation.
    \item \textbf{Naive sequential (lower target):} A single model is trained on each data subset incrementally, with each stage initialized from the previous stage's final checkpoint. This naive continual learning approach is susceptible to both catastrophic forgetting and the lack of co-observation. 
\end{itemize}
Because the ensemble preserves every historical state of the model, features learned at any stage remain available to the linear probe and its forgetting is mitigated by definition. 
Conversely, incremental joint training establishes an ``upper target'' that accounts for the loss of plasticity from the repeated optimization trajectory that the other sequential learning approaches are subject to. We included joint (offline) training to visualize that degradation. To the best of our knowledge, the ensemble serves as the most controlled approximation for a perfect-retention baseline to isolate the co-observation effect as cleanly as possible. Also, we do not aim to position the ensemble-baseline as a practical motivated choice of a continual learning approach, but use it as an analytical device.

\subsection{Controlling for representational capacity}
\label{sec:cntrl_rep_capacity}
A methodological challenge arises when comparing the ensemble to single-model baselines. First, the concatenated ensemble yields an expanded representation of dimension $T \times d$. To ensure a rigorous comparison of feature quality rather than raw capacity, we apply PCA where appropriate, retaining the $d$ directions of greatest variance. This projection does not perfectly simulate a single model that has mitigated forgetting, but corrects for differences due to architectural bloat.
Second, while our ensemble of frozen checkpoints prevents the loss of previously acquired information, i.e. forgetting, it inherently differs from a hypothetical ``ideal'' single model. Specifically, it does not fully account for the seamless consolidation of knowledge via forward and backward transfer. To gauge the influence of this potential confound regarding our experimental setup (Section~\ref{sec:coob_in_cl}), we empirically ablate it in Appendix~\ref{appx:emp_verification_ensemble} and find it to be minor.

\section{Proof of principle}
To provide a proof-of-principle for the co-observation effect, we design an experiment where we can explicitly control the available information at each training stage. This allows us to trace how separate observation, in comparison to joint observation, affects the learned representation.

\begin{figure}[!ht]
    \centering
    \includegraphics[width=0.98\linewidth]{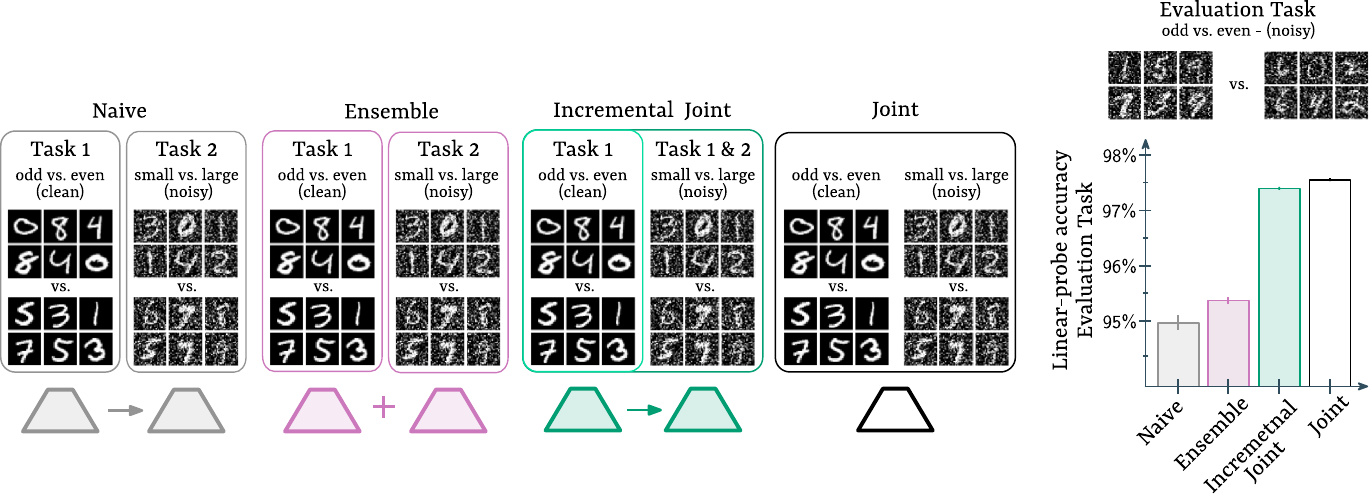}
    \caption{\textbf{Proof-of-principle MNIST experiment, demonstrating the co-observation effect.} Four approaches are compared: naive sequential training, ensemble, incremental joint, and (offline) joint training. Task~1 is binary classification of odd vs.\ even digits on clean images. Task~2 is binary classification of small ($\leq 4$) vs.\ large ($\geq 5$) digits on images augmented with Gaussian noise. The evaluation task is odd vs.\ even classification on images that exhibit similar Gaussian noise as Task~2. Both joint training variants achieve higher LP-accuracy, demonstrating that noise features transfer to the ``odd vs. even"-evaluation task when noisy data is being observed simultaneously, even when associated with a different task. This is contrasted by the ensemble, which preserves all learned representations but still achieves lower LP-accuracy. The gap between ensemble and joint training we attribute to the co-observation effect.
    \vspace{2em}
    }
    \label{fig:mnist_exp}
\end{figure}

We construct a sequence of two binary classification tasks from MNIST digits~\citep{lecun1998gradient}. Task 1 ($\mathcal{D}_1$) is odd vs. even classification on clean images. Task 2 ($\mathcal{D}_2$) is small (digits 0–4) vs. large (digits 5–9) classification on images corrupted with Gaussian noise (severity level 5, applied via the \textit{imgaug} library~\citep{imgaug}). We select the evaluation task to cover odd vs. even classification on noisy images (Figure~\ref{fig:mnist_exp}). 
Success on this evaluation task requires combining elements from both training tasks, namely the decision boundary from Task~1 and the ability to handle noisy inputs as encountered in Task~2. The question is whether the features, learned in the context of Task~1, remain available and transferable when learned sequentially versus jointly. For the experiments in this section, we utilize a 3-layer MLP backbone with $400$ hidden units per layer. The models are trained for $10$ epochs per task using a batch size of $64$. Optimization is performed using Stochastic Gradient Descent (SGD) with a peak learning rate of $0.01$, momentum of $0.9$, and gradient clipping at $1.0$, regulated by a cosine annealing schedule (5\% warm-up phase). We report the average performance across three independent runs with varying random seeds.

The results, displayed in Figure~\ref{fig:mnist_exp}, show that the ensemble achieves higher LP-accuracy than naive sequential training, presenting the influence of forgetting. However, incremental joint training further improves accuracy over the ensemble, pointing to another source of knowledge gain that is not explained by mitigating forgetting, as the ensemble preserves all past knowledge, but by co-observation that distinguishes joint training and separate training. To elaborate, in separate training, the model learns features that are sufficient for the current task's data, but in a form that generalizes primarily within that training context. As such, Task~2 does expose the model to noise and the resulting representation can handle noisy inputs for the small/large classification, but this capability does not transfer to the odd/even task when trained sequentially. In contrast, joint training, i.e.\ simultaneous observation, allows this transfer of knowledge. When data of both tasks are observed together during training, features that serve both objectives are realized.
\ \\

% Figure: MNIST Experiment

% \newpage
\section{Co-observation in continual learning}
\label{sec:coob_in_cl}
While the controlled experiment in the previous section showcases the effect of co-observation in a controlled but artificially designed scenario, we now continue to demonstrate its influence as a generic property to be taken into account in sequential learning. Continual learning benchmarks typically induce a severe distribution shift between stages (e.g., class-incremental learning). To emphasize the generality of our findings beyond artifacts of such amplified distribution shift or class imbalance, we evaluate the co-observation gap in an environment of uniformly partitioned incremental data. Following \citet{lee2023chunking}, we employ a ``chunking'' protocol (data-incremental learning) where the global training dataset is randomly and uniformly partitioned into four identically distributed chunks ($\mathcal{D}_1, \dots, \mathcal{D}_4$). In this setting, even without macroscopic distribution shifts between the sequential stages, we are able to observe representation deficits that can be directly attributed to the lack of co-observation.

We first assess the influence of co-observation across both supervised and self-supervised paradigms by comparing our diagnostic baselines. Subsequently, we evaluate exemplar replay and knowledge distillation in light of the co-observation effect. To ensure reproducibility, full details on the experimental setup are provided in Appendix~\ref{appx:detailed_exp_setup}. Further implementation specifics regarding the diagnostic ensemble and linear evaluation are detailed in Appendix~\ref{app:ensemble}, while the configurations for the continual learning methods are outlined in Appendix~\ref{app:cl_mechanisms}. Finally, comprehensive tables containing the numerical results for all experiments are available in Appendix~\ref{app:numerical_results}.

\subsection{Supervised and self-supervised continual learning}
To establish the co-observation gap as a generic property of sequential optimization, we first compare our four baselines: naive sequential training, the ensemble, incremental joint, and (offline) joint. As explained in Section~\ref{sec:cntrl_rep_capacity}, to control for the increased representational capacity of the ensemble, we also report the ensemble baseline with original backbone dimension $d$ reduced by PCA. We evaluate these baselines across both supervised and self-supervised paradigms using the $4$-chunk partitioned setup. In particular, we evaluate on CIFAR-100~\citep{krizhevsky2009learning} and ImageNet-100~\citep{russakovsky2015imagenet} (IN-100) using a ResNet-18~\citep{he2016deep} backbone for supervised learning and Barlow Twins~\citep{zbontar2021barlow}, and a ViT-B/16~\citep{dosovitskiy2021image} for I-JEPA~\citep{assran2023self}. 
All models are trained from scratch. For fair comparison, all training is constrained to an identical, fixed iteration budget per chunk. This fixed number of optimization steps is calibrated to match the (offline) joint training budget of $100$ epochs on CIFAR-100 and $90$ epochs on IN-100. Similarly, for the self-supervised methods (Barlow Twins and I-JEPA), the fixed iteration budget is calibrated to match $400$ epochs, aligned with \citet{fini2022self}.

\begin{figure}[!ht]
    \newlength{\figH}
    \setlength{\figH}{3.9cm}
    \centering
    \begin{subfigure}[t]{0.41\textwidth}
        \centering
        \includegraphics[height=\figH]{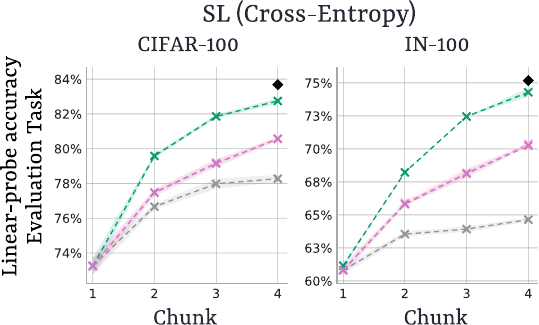}
        \caption{}
        \label{fig:main:left}
    \end{subfigure}
    \hfill
    \begin{subfigure}[t]{0.41\textwidth}
        \centering
        \includegraphics[height=\figH]{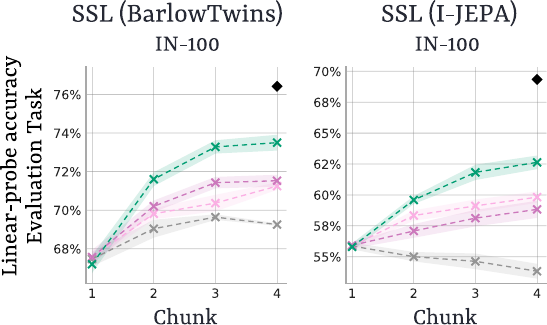}
        \caption{}
        \label{fig:main:right}
    \end{subfigure}
    \hfill
    \begin{subfigure}[t]{0.15\textwidth}
        \centering
        %\vspace{-2.5cm}
        \includegraphics[width=\textwidth]{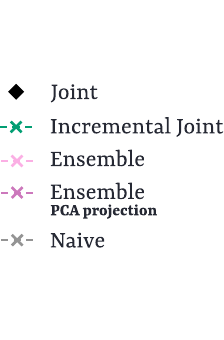}
    \end{subfigure}
    \vspace{-0.5em} % since we are not using subcaptions
    \caption{%
        \textbf{The co-observation effect across supervised and self-supervised learning paradigms.}
        Linear probing accuracy is reported across 4 sequential data chunks, showing the mean and standard error over 3 seeds (6 for I-JEPA).
        \textbf{(a)} Supervised learning (Cross-Entropy) on CIFAR-100 and ImageNet-100 (IN-100). \textbf{(b)} Self-supervised learning using Barlow Twins and I-JEPA on IN-100. In all settings, the beneficial effect of co-observation is clearly visible. The ensemble baseline prevents the forgetting seen in the Naive baseline, and consistently falls short of the incremental joint training. This residual difference between ensemble (separate training) and incremental joint highlights the impact of co-observation, demonstrating that simultaneously observing data is important for representation learning, regardless of the objective function.
    }
    \label{fig:main}
\end{figure}

\textbf{Supervised continual learning:}
As illustrated in Figure~\ref{fig:main:left}, and further extended in Appendix~\ref{appx:more_chunking}, naive sequential training exhibits the lowest linear probing accuracy, affirming the impact of catastrophic forgetting on the representation even in the absence of explicit class-incremental distribution shifts. The  ensemble baseline significantly improves upon naive sequential training and is not affected when controlling for representational capacity by PCA reduction, indicated by both lines overlapping. %
A clear benefit is visible from co-observation as the incremental joint training consistently improves over the ensemble. 
Because the ensemble retains all past representations, this residual difference highlights the co-observation effect, i.e. the representational benefit derived from simultaneously observing data across different chunks, is a property that is fundamentally lost in separate training.

\textbf{Self-Supervised continual learning:}
To investigate whether the co-observation effect is specific to supervised cross-entropy optimization, we extend our evaluation to self-supervised learning (SSL). Unlike supervised learning, which leverages explicit class boundaries, SSL methods rely on inferring diverse features within the training batches to form robust, invariant representations, which are thought to be more robust to forgetting~\citep{fini2022self}. We evaluate two distinct families of SSL architectures on IN-100: Barlow Twins \citep{zbontar2021barlow}, which relies on redundancy reduction, and I-JEPA \citep{assran2023self}, a recent joint-embedding predictive framework. 

The results (Figure~\ref{fig:main:right}) demonstrate that the co-observation gap is also pronounced in the self-supervised regime. Interestingly, applying PCA reduction to the ensemble yields varying effects depending on the SSL objective. For Barlow Twins it can occasionally improve the representation quality, while degrading it for I-JEPA. Yet, across both SSL algorithms and regardless of the capacity control, our ensemble baselines consistently evaluate lower than the representation quality obtained from the incremental joint model. This result indicates that the co-observation gap is a consequence of separate representation learning: synergistic features that span across data partitions cannot be fully realized without simultaneous observation, regardless of the underlying optimization objective.
While it is difficult to perfectly isolate the co-observation effect from all sequential optimization dynamics, we account for the primary alternative explanation of a deficit in knowledge transfer in Appendix~\ref{appx:emp_verification_ensemble}. Supported by our controls, and this ablation, the persistent representational deficit observed across both supervised and distinct self-supervised paradigms strongly points to the structural lack of simultaneous data observation as the primary driver.

\subsection{Relation to continual learning mechanisms}
Having confirmed the co-observation gap in generic continual learning scenarios, we now examine how replay and knowledge distillation compare to our established baselines, to decouple their ability to incrementally improve generalization in the model's representation relative to the mitigation of forgetting alone. 

\begin{figure}[!b]
    % \newlength{\figH}
    \setlength{\figH}{3.7cm}
    \centering
    \begin{subfigure}[t]{0.39\textwidth}
        \centering
        \includegraphics[height=\figH]{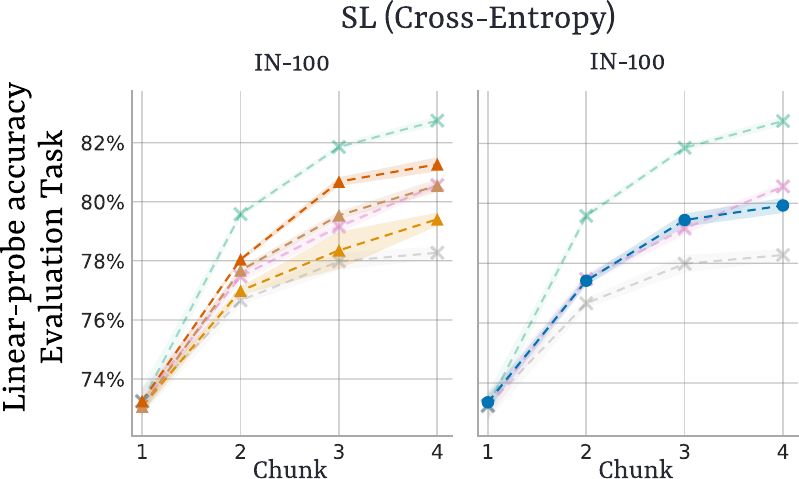}\\[0.7cm]
        \caption{}
        \label{fig:replay_and_distill:left}
    \end{subfigure}
    \hfill
    \begin{subfigure}[t]{0.6\textwidth}
        \centering
        \includegraphics[height=\figH]{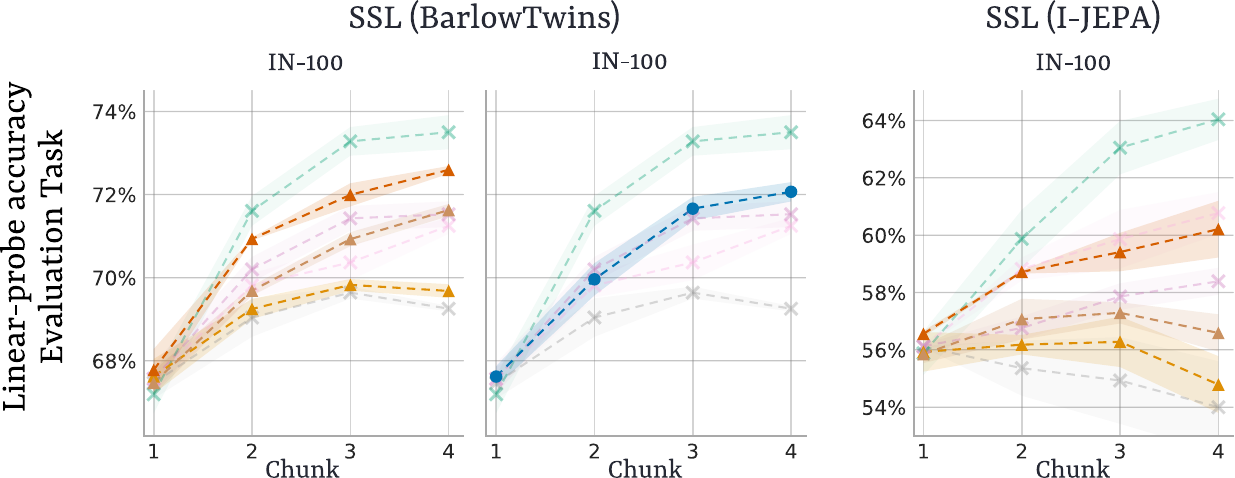}\\[0.7cm]
        \caption{}
        \label{fig:replay_and_distill:right}
    \end{subfigure}
    \hfill
    \ \\[-1cm]
    \includegraphics[width=0.9\textwidth]{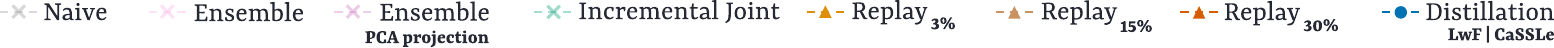}\\[0.4cm]
    \caption{%
        \textbf{Interaction of continual learning mechanisms with the co-observation gap.} %
        %Linear probing accuracy evaluated across four sequential data chunks 
        Linear probing accuracy is reported across 4 sequential data chunks, showing the mean and standard error over 3 seeds
        for \textbf{(a)} Supervised Learning and \textbf{(b)} Self-Supervised Learning (Barlow Twins and I-JEPA). Comparing experience replay and knowledge distillation (LwF for SL; CaSSLe for Barlow) against the naive-, joint-incremental-, and ensemble-baselines, empirically demonstrates the relationship of both methods to mitigation of forgetting and obtained generalization from the representation learning perspective.
    }
    \label{fig:replay_and_distill}
\end{figure}

\textbf{Experience Replay:}
Replaying stored data is intuitively the most direct approach to bridging separate learning stages. Mechanistically, it creates a minibatch-level approximation of the joint data distribution by presenting previous and new data together. While replay is predominantly renowned as a forgetting-mitigation tool, we evaluate it here to explicitly contextualize this minibatch approximation as mechanism that enables data co-observation.
We evaluate replay on IN-100 with buffers allowing the storage of $1250 (\sim3\%)$, $6250 (\sim15\%)$, and $12500 (\sim30\%)$ exemplars of the data from each previously observed chunk, where new data and all data from the replay buffers are combined and sampled i.i.d. into mini-batches during optimization~\citep{rebuffi2017icarl,wu2019large,cho2026forget}.

In terms of raw values, as expected, the results presented in Figure~\ref{fig:replay_and_distill} show %little surprisingly
that larger replay buffers yield higher accuracy. However, our interest is in the comparison to our baselines which allows to reason about a more nuanced dynamic. 
While sparse rehearsal (e.g., $3\%$) has been shown to yield improvements when evaluated via end-to-end accuracy~\citep{chaudhry2019tiny}, our linear-probing evaluation reveals a different dynamic at representation level. Here, replay with tiny buffers struggles to consistently match the diagnostic ensemble. Storing approximately $15\%$ of the data appears sufficient for the replay mechanism to effectively match the performance of the ensemble. When the buffer size is increased to $30\%$ of the observed exemplars per stage, the linear probing accuracy pushes past the ensemble baseline, closing half the gap to incremental joint training. These results hold for supervised and self-supervised training with Barlow Twins and I-JEPA.

The replay mechanism inherently conflates the mitigation of forgetting with the benefits of co-observation. However, our comparison against our ensemble baseline yields two crucial takeaways. First, our findings suggest that replay's effectiveness is not bound merely by mitigating forgetting, but by actively reintroducing data co-observation. Because we strictly control the training budget across all setups, the gains from larger replay buffers are not an artifact of extended optimization on more data. Rather, the improved approximation of the joint distribution at the mini-batch level -- instantiating the mechanism underlying co-observation -- drives altered optimization trajectories that allow the model to synthesize cross-distribution features that perfect retention alone cannot recover. Second, there is a significant amount of stored exemplars necessary to improve beyond results achievable by mitigation of forgetting, although the specific number of samples required will inherently vary based on the domain and dataset diversity.

\textbf{Distillation:}
As a popular memory-free alternative to rehearsal, we evaluate knowledge distillation. We apply Learning without Forgetting (LwF) \citep{li2016learning} for the supervised setup, and CaSSLe \citep{fini2022self} for the self-supervised Barlow Twins setup. 

As shown in Figure~\ref{fig:replay_and_distill}, distillation achieves linear probing accuracies that closely mirror those of our ensemble and the $15\%$ replay buffer. Knowledge distillation has previously been shown to be an effective retention mechanism, successfully preventing representation-level forgetting~\citep{ni2023representation}. We deem it important to emphasize that knowledge distillation here is not failing to prevent forgetting, as indicated by its parity with the ensemble. Rather, distillation from the ``incomplete" representation of the previous model does not facilitate the co-observation of information that would be required to recover the representational effectiveness found in large-scale replay.

\section{Related Work}

\subsection{Forgetting, plasticity, and a third factor}
The continual learning literature has predominantly framed the challenges of sequential optimization along the stability-plasticity axis~\citep{mermillod2013stability}, typically addressed through mechanisms such as regularization~\citep{kirkpatrick2017overcoming, zenke2017continual, li2016learning}, replay~\citep{rebuffi2017icarl, lopezpaz2017gradient, buzzega2020dark, rolnick2019experience}, or architecture expansion~\citep{rusu2016progressive, yan2021der}. Progress is measured against a joint-training upper target, implicitly assuming that a model with perfect stability (zero forgetting) and perfect plasticity (ability to incorporate new information) would naturally recover joint-training performance~\citep{wang2024comprehensive, momeni2025achieving}. While recent work by \citet{dohare2024loss} and others~\citep{verwimp2025same, cho2026forget} has established \emph{loss of plasticity} as a distinct degradation independent of forgetting, we identify the co-observation effect as a third factor. 

Several recent studies have empirically encountered symptoms of the co-observation effect, though these observations have typically been isolated within specific continual learning setups: \textbf{In class-incremental scenarios}, \citet{ni2023representation} observed ``inter-phase confusion'' persisting after addressing forgetting, and \citet{soutif2021importance} noted a deficit in cross-task features. 
While conceptually related, these observations have been confined to class-incremental setups; we demonstrate that they are part of a general deficit in continual representation learning, extending well beyond class-incremental considerations.
\textbf{Moving beyond class-incremental boundaries}, \citet{busch2025where} demonstrated that ``continual confounders'' prevent sequential learners from discovering generalizable rules. To clearly isolate this effect, the authors systematically introduce explicitly constructed, adversarial spurious correlations within a synthetic dataset. Our work highlights the broader implications of this result for separate training, demonstrating that such confounding effects can be considered to arise naturally simply from the sequence of the data itself. To do so, we build on the ``chunking" scenario previously utilized by \citet{lee2023chunking} to relate the impact of separate training to \emph{forgetting}. 
We shift the focus to \emph{data co-observation}, demonstrating that the co-observation effect emerges generically from random partitioning and affects generalization alongside forgetting, class imbalance, or adversarial structures.

Finally, our work establishes that this effect on continual representation learning is not merely an artifact of supervised training dynamics or the narrow decision boundaries induced by cross-entropy. Self-supervised learning (SSL) methods are broadly considered to learn more diverse, task-agnostic features~\citep{fini2022self, madaan2022representational}. Despite these advantages, we show that the lack of co-observation affects learning similarly across the distinct SSL frameworks of Barlow Twins~\citep{zbontar2021barlow} and I-JEPA~\citep{assran2023self}. This establishes the co-observation effect as a fundamental limitation of fragmented optimization, affecting both supervised and self-supervised sequential learning.

\subsection{Representation evaluation and the ensemble baseline}
End-to-end accuracy in continual learning often conflates actual representation quality with the dynamics of a continually updated classifier head~\citep{wu2019large}. To address this, \citet{davari2022probing} used linear probing (LP)~\citep{alain2016understanding} as an evaluation tool for CL, demonstrating that representation-level forgetting under naive finetuning is substantially milder than end-to-end metrics often suggest. 
We also adopt LP to evaluate the knowledge accumulated within our model at representation level. Thereby, our approach is more closely related to \citet{hess2023knowledge}, who combined LP evaluation with an ensemble baseline to demonstrate that even seemingly ``mild" representation-level forgetting in absolute numbers still exerts a detrimental effect on the effectiveness of continual representation learning. Instead of utilizing probing to highlight the penalty of forgetting, in the current work, we leverage the ensemble as a baseline to control for forgetting, enabling our novel analysis regarding the isolation of the co-observation effect.

\subsection{Multi-task learning and joint optimization}
Historically, continual learning shares conceptual roots with multi-task learning~(MTL). Both paradigms strive to accumulate and leverage knowledge across multiple objectives \citep{chen2018lifelong} and early continual learning works were differentiated from MTL by training the tasks incrementally instead of jointly. %In MTL, the goal is to obtain superior shared representations by observing related tasks jointly. 
Regarding MTL, \citet{caruana1997multitask} showed that learning tasks in parallel can improve generalization because related tasks provide a mutually beneficial inductive bias, helping the model discover features it would not find when trained on either task in isolation. \citet{ruder2017overview} articulated several mechanisms driving this benefit, including implicit data augmentation, attention focusing, and regularization through multi-objective optimization. \citet{baxter2000model} provided a theoretical foundation for this, showing that the risk of overfitting shared parameters decreases with the number of co-observed tasks. Furthermore, as demonstrated by \citet{bengio2009curriculum}, the specific composition of training data significantly influences optimization trajectories in non-convex landscapes. These works underscore that \emph{which} data is observed \emph{simultaneously}, fundamentally shapes the quality of the learned solutions.

To the best of our knowledge the classical MTL desideratum of improving representation quality was largely left aside in continual learning, presumably because of the observed overwhelming impact of catastrophic forgetting. In this sense, our work can be understood as demonstrating that even within a single task (rather than multiple different tasks), fragmenting data into separate portions leads to reduced generalization of the model.
Consequently, a limit is placed on the representational quality that can be achieved from strict separate learning, regardless of how well the aspect of forgetting is controlled.

\section{Discussion}
\subsection{Limiting factors other than forgetting}
Continual learning algorithms are expected to improve over naive sequential training. However, one could argue that there are cases where reaching the joint training upper target is a rather theoretical construct because of strict limitations in compute~\citep{prabhu2023computationally,verwimp2024continual}, memory~\citep{de2021continual, chaudhry2019tiny}, privacy boundaries~\citep{farquhar2019differentially}, the need to operate in an open world~\citep{mundt2023wholistic}, etc. Nevertheless, achieving best possible results is the goal and a prevailing narrative heavily focuses on catastrophic forgetting as the primary culprit for the performance drop observed when training continually, with some recent pushes to also consider effects from plasticity losses~\citep{cho2026forget, verwimp2025same}. Our work points out yet another, fundamental, effect where previous knowledge is perfectly retained or protected, but the model's generalization falls short of the upper target. That is, without requiring any particularity of the distribution shift underlying the separate observations, apart from the samples being disjoint. Different instances of this effect have been studied, predominantly with respect to class-incremental learning, where intuitively observing instances of different classes together is required to construct discriminating features~\citep{wu2019large, soutif2021importance, ni2023representation}. But also regarding the potential of confounding factors in the separately observed training data, which inflict deficiencies on the continually training model that are not easy to recover~\citep{busch2025where}. We demonstrate data co-observation as an inherent and distinct limiting factor in separate training, without additional prerequisites as studied in prior works.

\subsection{Connection to large pre-trained models and LLMs}
Recent continual learning literature increasingly leverages pre-trained models. Approaches that leave the backbone frozen reduce continual learning to incremental prototype memorization \citep{janson2022}. In such settings, the discovery of cross-distribution synergies is typically of minor importance and the co-observation effect is bypassed. Methods employing conservative fine-tuning, e.g. applying slow learning rates \citep{zhang2023slca} or utilizing adapters \citep{aghasanli2025prototype}, aim to make only slight adjustments to the pre-trained features. Arguably, the magnitude of new learning is inherently small relative to the initial pre-training phase, and optimization is heavily regularized to protect existing weights. Consequently, the dynamics of catastrophic forgetting and its mitigation largely overpower any representational gains that would theoretically emerge from co-observation. 
Continual pre-training to broadly update and improve the generalization of the backbone in a more foundational manner \citep{hu2022well} is where we hypothesize co-observation effects to be more important.

While our empirical analysis is conducted on standard vision benchmarks, evidence from the large language model~(LLM) literature suggests that the interplay between data co-occurrence and representation quality extends to larger scales. \citet{ye2024data} demonstrate that the \emph{composition} of training batches during LLM pre-training, specifically, cross-domain interaction effects, quantitatively predicts model performance. Since their analysis concerns standard i.i.d.\ pre-training rather than sequential learning, this provides evidence that which data is observed together shapes learned representations independently of any forgetting dynamics. In the continual pre-training setting, \citet{ibrahim2024simple} show that replaying previous data into new training batches is necessary to match the performance of retraining from scratch, a gap that learning rate adjustments alone cannot close. While this deficit is conventionally attributed to forgetting, our framework raises the complementary hypothesis that replay also restores co-observation conditions that enable the model to discover cross-distribution patterns. Disentangling these two roles of replay, forgetting prevention versus co-observation restoration, at LLM scale constitutes a natural direction of investigation. The diagnostic framework introduced in this work, comparing forgetting-free ensembles against joint training via linear probing, offers a principled starting point for this investigation.

\section{Conclusion}
We empirically separated the effect of data co-observation from stability-plasticity considerations for the representation-level generalization of continually trained deep neural networks, and demonstrated the generic emergence of that effect across supervised and self-supervised training in data-incremental scenarios.
While the representational synergies of joint training are well-established in the multi-task learning literature, we demonstrate such a performance gap persists generically in standard vision benchmarks even in ``chunking'' scenarios.
We argue that making the influence of co-observation explicit and positioning it distinct from forgetting and plasticity, as a third facet of continual learning, is important for the %continual learning 
community: assuming plasticity and successful forgetting mitigation whilst learning without co-observation appears to hit a performance ceiling that can be significantly lower than the upper target of a jointly trained model. 
We emphasize that we do not claim the performance penalty of co-observation universally rivals the sheer magnitude of catastrophic forgetting. As discussed in the context of pre-trained models, forgetting often remains the most striking source of degradation in many applications. Nevertheless, we deem explicitly isolating and understanding the co-observation effect to be conceptually vital for mapping the full dynamics and limits of continual learning.
Improvement beyond requires continual learning algorithms to actively synthesize cross-data synergies. How to best facilitate such synergies, especially in compute-bounded settings, needs to be the subject of further research.

\section{Limitations}
We see three limitations to this work. The first regards our diagnostic methodology, which intends to isolate the co-observation gap using linear probing and an ensemble baseline. While this effectively removes confounders like classifier deviation and forgetting, strictly controlling the ensemble's capacity requires post-hoc dimensionality reduction (in our case using PCA). Although an ideal baseline would natively match the original model's parameter count, we deem our capacity-controlled ensemble to serve as the most viable analytical proxy. 
Secondly, we employed uniform optimization schedules to ensure fair comparability between joint and separate training, rather than to maximize absolute performance. Other continual optimizers might yield higher accuracies, mitigate part of the forgetting or better control plasticity, but we expect the underlying trends due to data co-observation to remain robust. Finally, while we demonstrate the effect from co-observation across supervised and self-supervised paradigms on moderate-scale vision tasks, representational dynamics can shift at massive scales. Future work must investigate whether billion-parameter over-parameterization naturally mitigates or exacerbates the need for simultaneous data observation.

\newpage
\section*{Acknowledgments}
This paper is part of a project that has received funding from the European Union under the Horizon 2020 research and innovation program (ERC project KeepOnLearning, grant agreement No. 101021347), and KU Leuven C1 project (Reference: C14/23/100). Part of this work was conducted while Abhishek was affiliated with ESAT-PSI, KU Leuven, Belgium. The computational resources and services used in this work were partly provided by the VSC (Flemish Supercomputer Center). 

%\newpage
\bibliography{collas2026_conference}
\bibliographystyle{collas2026_conference}

\newpage
\appendix
\section*{Appendix}
The following appendix provides supplementary details regarding our experimental setups, further details on our evaluation methodology, and quantitative results. The contents are structured as follows:

\begin{itemize}
    \item \textbf{Appendix~\ref{appx:detailed_exp_setup}} details the optimization hyperparameters, dataset preprocessing, and augmentation pipelines for both the supervised and self-supervised pre-training regimes.
    \item \textbf{Appendix~\ref{app:ensemble}} expands upon our evaluation framework, describing the construction of the perfect-retention ensemble, the capacity-controlling dimensionality reduction via PCA, and the linear probing protocol, in full detail.
    \item \textbf{Appendix~\ref{appx:emp_verification_ensemble}}  provides empirical verification of the diagnostic ensemble, demonstrating its zero-forgetting properties on previously observed training data and ablating the potential confounding influence of forward and backward transfer using distillation-enhanced checkpoints.
    \item \textbf{Appendix~\ref{app:cl_mechanisms}} formally defines the continual learning mechanisms evaluated in this work, specifically our adaptations of Experience Replay and Knowledge Distillation (LwF and CaSSLe).
    \item \textbf{Appendix~\ref{appx:more_chunking}} presents additional results extending the supervised learning evaluation on ImageNet-100 to longer continual sequences of $8$ and $12$ chunks, demonstrating the robustness of the co-observation gap under increased data fragmentation.
    \item \textbf{Appendix~\ref{app:numerical_results}} provides the comprehensive numerical results, corresponding to the empirical evaluations and figures presented in the main text Section~\ref{sec:coob_in_cl}.
\end{itemize}

\section{Detailed Experimental Setup}
\label{appx:detailed_exp_setup}
\textbf{Datasets and Augmentations.} 
For CIFAR-100, training images (original resolution $32 \times 32$) are augmented using random cropping (padding of $4$, reflect mode) and random horizontal flipping ($p=0.5$). Images are normalized with $\mu = (0.485, 0.456, 0.406)$ and $\sigma = (0.229, 0.224, 0.225)$. At evaluation, images are strictly evaluated at $32 \times 32$.

For ImageNet-100 (IN-100), images are stored at $256$ pixels on their longer side. During training, we apply a RandomResizedCrop to $224 \times 224$, a random horizontal flip, and RandAugment ($2$ operations, magnitude $9$). At evaluation, images are center cropped to $224 \times 224$, normalized with $\mu = (0.4914, 0.4822, 0.4465)$ and $\sigma = (0.2023, 0.1994, 0.2010)$.

\textbf{Supervised Learning Configurations.}
For supervised training, we utilize a ResNet-18 backbone. For CIFAR-100, the backbone is adjusted to accommodate the smaller input size, as proposed by \citet{he2016deep} (i.e., replacing the initial $7 \times 7$ convolution with a $3 \times 3$ convolution and removing the first max-pooling layer). 

The model is optimized using SGD with a momentum of $0.9$, a weight decay of $5 \times 10^{-4}$, and gradient clipping set to $1.0$. The learning rate follows a warmup phase constituting $5\%$ of the total iterations, followed by a cosine annealing schedule with a peak learning rate of $0.1$, and a final learning rate of $10^{-5}$. To maintain consistent training dynamics across datasets, the total number of iterations is scaled to be equivalent to approximately $100$ epochs for CIFAR-100 and $90$ epochs for IN-100 on the respective full datasets. Specifically, training on CIFAR-100 proceeds for $9,750$ iterations per chunk with a mini-batch size of $128$, while training on IN-100 proceeds for $11,070$ iterations per chunk with a mini-batch size of $256$. 

The linear probe is trained mirroring the linear evaluation protocol of \citet{fini2022self}. We use a fixed number of $100$ epochs on the (optionally reduced) frozen representations, using SGD with a momentum of $0.9$, an initial learning rate of $0.1$, and no weight decay. The learning rate is decayed by a factor of $10$ at $60$ and $80$ epochs.

\textbf{Self-Supervised Learning Configurations.}
For the self-supervised continual learning experiments, we evaluate two distinct frameworks with their respective standard configurations. 

\textbf{Barlow Twins.} The ResNet-18 encoder is paired with a $3$-layer MLP projector (hidden and output dimensions of $2048$) using BatchNorm and ReLU activations. The objective uses an off-diagonal scaling factor $\lambda = 5 \times 10^{-3}$ and an overall loss scale of $0.1$. Two views are generated per image using random resized cropping (scale $[0.08, 1.0]$), horizontal flipping, color jitter, Gaussian blur, and random solarization. The model is optimized using LARS (learning rate $0.4$, weight decay $10^{-4}$, momentum $0.9$) with a warmup-cosine schedule for $400$ epochs per chunk and a batch size of $256$.

\textbf{I-JEPA.} The ViT-B/16 encoder (patch size 16, $224 \times 224$ input) is trained alongside an Exponential Moving Average (EMA) target encoder, with the momentum coefficient annealed from $0.996$ to $1.0$. The predictor is a depth-6 ViT with an embedding dimension of $384$. A block-masking strategy samples one context mask ($85\text{--}100\%$ of patches) and four target masks ($15\text{--}20\%$ of patches). The model minimizes the smooth-$\ell_1$ (Huber) loss and is optimized using Adam (learning rate $10^{-3}$, weight decay $0.04$) with a warmup-cosine schedule for $400$ epochs per chunk at a batch size of $512$.

\section{Method details: Ensemble}
\label{app:ensemble}

\subsection{Ensemble Construction}
The ensemble-baseline introduced in Section~\ref{sec:isolating_coob} is designed to provide a perfect-retention baseline by storing the exact state of the continually training model after each training phase. After completing training on experience $t$, a frozen copy of the feature extractor (backbone) $f_{\theta_t}$ is appended to a persistent buffer $\mathcal{B} = \{f_{\theta_1}, f_{\theta_2}, \ldots, f_{\theta_t}\}$. Thereafter, these backbone weights are never updated. 

During evaluation, an input $\mathbf{x}$ is passed through all $|\mathcal{B}|$ frozen backbones, as illustrated in Figure~\ref{fig:ensemble_explain}. The resulting features are concatenated along the channel dimension to form a single combined representation $\mathbf{z} \in \mathbb{R}^{|\mathcal{B}| \cdot d}$, where $d$ is the feature dimension of a single backbone. This concatenated representation is then used for downstream linear probing.

\subsection{Dimensionality Reduction}
\label{app:pca_reduction}
Because the ensemble concatenates representations from $|\mathcal{B}|$ checkpoints, its output dimension $|\mathcal{B}| \cdot d$ scales linearly with the number of training phases. To ensure a fair comparison against single-model baselines and isolate feature quality from increased dimensional capacity, we apply an optional dimensionality reduction step to compress $\mathbf{z}$ back to a target dimension $d' = d$ before linear probing. For the reduction a PCA is fitted exclusively on the concatenated representations of the downstream training set, retaining the $d'$ components with the largest variance. The fitted projection matrix is then fixed and applied to the evaluation set to prevent information leakage. This reduction step is only applied when evaluating the ensemble. 

\subsection{Linear-Probe Evaluation}
\label{app:linear_probe}
At each evaluation point, a newly initialized linear classification head $\mathbf{W} \in \mathbb{R}^{C \times d_{\text{in}}}$ is instantiated and trained on the downstream training set, where $C$ is the number of downstream classes and $d_{\text{in}}$ is the appropriate feature dimension.

To optimize training efficiency, all training-set inputs are passed through the frozen backbone(s) once, and the resulting representations are cached. The linear probe is then trained on these cached features using a standard cross-entropy loss. Similarly to \citet{fini2022self} the optimizer is an SGD (with momentum and either a multi-step or cosine annealing learning rate schedule). After training, the probe is evaluated on the downstream test set via a single forward pass, reporting top-$1$ accuracy.

\section{Empirical verification of knowledge retention in the ensemble}
\label{appx:emp_verification_ensemble}
Our diagnostic ensemble serves as analytical device to isolate the co-observation effect from the standard stability-plasticity considerations as well as possible. To ensure the validity of this isolation, this section provides empirical verification that the ensemble reliably controls for catastrophic forgetting (perfect retention), and ablates potential confounding deficits regarding knowledge transfer.

\subsection{Empirical verification of perfect retention}
\label{appx:emp_verification_ensemble_1}
In Section~\ref{subsec:three_baselines}, we introduce an ensemble baseline designed as an analytical device to achieve perfect retention (zero forgetting). To empirically corroborate that our ensemble baseline exhibits this stability, we mirror the diagnostic methodology utilized by \citet{lee2023chunking} and conduct an evaluation directly on the training data. Because data-incremental ``chunking'' scenarios do not inherently possess a separate validation set per chunk, evaluating performance on the previously seen training data serves as a direct proxy for information retention. 
\begin{figure}[!ht]
    \centering
    \includegraphics[width=0.85\linewidth]{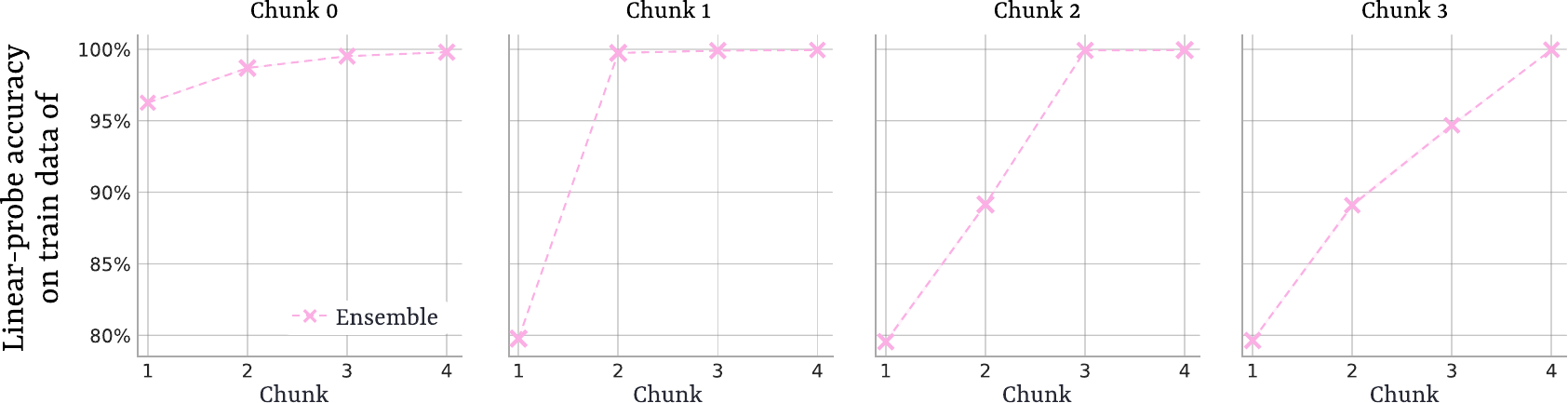}
    \caption{Empirical verification of zero forgetting in the diagnostic ensemble baseline. The linear probe accuracy is evaluated on the training sets of individual chunks ($T_0$ through $T_3$) across 4 sequential training stages. After a chunk is actively trained on, its train-set accuracy does not decrease in subsequent stages.}
    \label{fig:no_forget}
\end{figure}

As demonstrated in Figure~\ref{fig:no_forget}, once the ensemble trains on a given chunk, its accuracy on that chunk's training data never declines in any subsequent stage. Instead, we observe an increase in accuracy for chunk $0$ and forward transfer w.r.t. later chunks. 
This occurs because the representations of past chunks are perfectly preserved, and concatenating newly frozen backbones from later stages only provides the global linear probe with strictly more features to combine, effectively enabling backward transfer at the readout level. This confirms the ensemble operates as a strict perfect-retention baseline.

\subsection{Ablation on knowledge transfer capabilities}
\label{appx:emp_verification_ensemble_2}
While the ensemble perfectly mitigates the loss of previously acquired information post-hoc, it is constructed from a naive sequential continual  learner. Consequently, the individual checkpoints inherently suffer from forgetting during the sequential training process itself, which can impair forward transfer (the ability of past learned features to facilitate future learning). This means the ensemble does not fully account for the seamless consolidation of knowledge via "ideal" forward and backward transfer that a perfect, single continually learning model would possess. 
To gauge this potential confounding influence, we conduct an ablation where we construct the ensemble using checkpoints from models equipped with stronger transfer and retention properties, rather than naive sequential checkpoints. Specifically, we utilize knowledge distillation methods, i.e. LwF~\citep{li2016learning} for the supervised setting, and CaSSLe~\citep{fini2022self} for the self-supervised setting. We choose distillation because it does not reintroduce data co-observation, preserving the validity of the ablation.

\begin{figure}[!ht]
    \centering
    \includegraphics[width=0.75\linewidth]{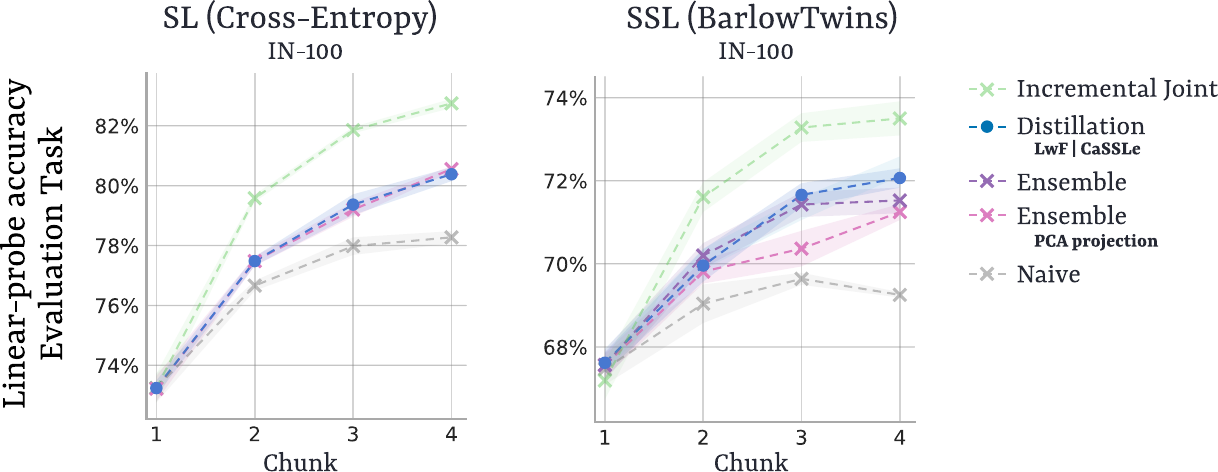}
    \caption{Ablation of forward/backward transfer using distillation-enhanced checkpoints. Comparing the standard diagnostic ensemble (built from naive sequential checkpoints) against an ensemble built from distillation-enhanced checkpoints (LwF/CaSSLe) across sequential chunks. Even with enhanced transfer properties, the ensemble at best barely surpasses the standard baseline and fails to close the gap to joint-incremental training.}
    \label{fig:distill_ens}
\end{figure}

As shown in Figure~\ref{fig:distill_ens}, substituting the naive checkpoints with distillation-enhanced checkpoints yields marginal (in case of CaSSLe) to no (in case of LwF) improvements. The performance of this enhanced ensemble tightly tracks the standard ensemble baseline and remains significantly below incremental joint training. Combined with our conceptual arguments in Section~\ref{subsec:conceptual_explanation}, this supports that the residual performance gap between the ensemble and joint training can be attributed to the representational deficit caused by the lack of simultaneous data co-observation, rather than being solely an artifact of poor transfer interaction between isolated checkpoints.

%

%\newpage
\section{Method details: Continual Learning Mechanisms}
\label{app:cl_mechanisms}

Implementation of the experiments builds on the Avalanche CL library~\citep{JMLR:v24:23-0130}\footnote{\url{https://github.com/ContinualAI/avalanche}}, the CaSSLe~\citep{fini2022self}\footnote{\url{https://github.com/DonkeyShot21/cassle}} framework for self-supervised continual learning, and the official codebase of I-JEPA~\citep{assran2023self}\footnote{\url{https://github.com/facebookresearch/ijepa}}.

\subsection{Experience Replay}
\label{app:replay}
Our implementation of experience replay maintains a persistent memory buffer $\mathcal{M}$ of past exemplars, with a buffer policy that stores $m$ samples per observed task. 
Similar to a number of works that utilize exemplar replay in the literature~\citep{rebuffi2017icarl, wu2019large, cho2026forget}, during optimization on experience $t$, the buffer $\mathcal{M}$ is concatenated directly onto the current-task dataset $\mathcal{D}_t$ to form a single combined dataset $\mathcal{D}_t^{+} = \mathcal{D}_t \cup \mathcal{M}$. Mini-batches of size $B$ are sampled uniformly from $\mathcal{D}_t^{+}$. Consequently, the expected fraction of buffer samples in any given training batch is given by $|\mathcal{M}| / (|\mathcal{D}_t| + |\mathcal{M}|)$. This approach naturally mirrors the data balance observed during cumulative joint training and automatically adapts as the buffer fills over successive experiences.

\subsection{Knowledge Distillation (Supervised: LwF)}
\label{app:lwf}
For knowledge distillation in the supervised continual learning case, we utilize Learning without Forgetting (LwF)~\cite{li2016learning}. After training on experience~$t$, a frozen copy of the model~$\theta^{t-1}$ is saved as the \emph{teacher}. When training on experience~$t+1$, the teacher's soft predictions over previously seen classes serve as targets for the current \emph{student} model.

Let $\mathbf{z}$ and $\mathbf{z}^{\text{prev}}$ denote the logits of the student and teacher models, respectively. The distillation loss for a previous task~$\tau$ is the Kullback-Leibler (KL) divergence between the softened distributions:
\begin{equation}
    \mathcal{L}_{\mathrm{KD}}^{\tau} = \mathrm{KL}\!\left(
        \sigma\!\left(\mathbf{z}^{\text{prev}}_{\mathcal{A}_\tau} / T\right)
        \;\Vert\;
        \sigma\!\left(\mathbf{z}_{\mathcal{A}_\tau} / T\right)
    \right),
\end{equation}
where $\sigma(\cdot)$ is the softmax function, $T = 2$ is the temperature scaling as proposed by the original authors, and $\mathcal{A}_{\tau}$ is the set of active class indices for task~$\tau$. The total loss is a weighted sum of the cross-entropy loss on the current experience and the distillation loss over all previously encountered tasks. To ensure stability, distillation is applied strictly over the output units $\mathcal{A}_\tau$ corresponding to classes the model has actively trained on.

\subsection{Knowledge Distillation (Self-Supervised: CaSSLe)}
\label{app:cassle}
To adapt distillation to the self-supervised regime, we employ the CaSSLe framework~\cite{fini2022self} using the Barlow Twins~\cite{zbontar2021barlow} objective. Barlow Twins learns representations by driving the empirical cross-correlation matrix $\mathbf{C}$ of two augmented views toward the identity matrix, utilizing an invariance term (diagonal) and a redundancy-reduction term (off-diagonal).

In the CaSSLe decorrelative variant, at the end of experience $t$, the encoder $f_{\tilde\theta}$ and projector $h_{\tilde\psi}$ are frozen to act as the teacher. During experience $t+1$, a lightweight, randomly initialized MLP predictor $g_\phi$ is added to the student model. The student processes the input through its encoder, projector, and the new predictor to yield predictions $\mathbf{P}$. The teacher processes the same input to yield target projections $\tilde{\mathbf{Z}}$. 

The distillation loss applies the Barlow Twins objective ($\mathcal{L}_{\mathrm{BT}}$) to align the student's predictions with the teacher's frozen projections:
\begin{equation}
    \mathcal{L}_{\mathrm{distill}} = \frac{1}{2}\Bigl[
        \mathcal{L}_{\mathrm{BT}}(\mathbf{P}^A,\, \tilde{\mathbf{Z}}^A) + \mathcal{L}_{\mathrm{BT}}(\mathbf{P}^B,\, \tilde{\mathbf{Z}}^B)
    \Bigr]
\end{equation}
The predictor $g_\phi$ is applied exclusively to the student network. This asymmetry serves two functions: it prevents representation collapse, and it decouples the student's current-task projector from the rigid distillation constraint, allowing the student to adapt to new data while the predictor bridges the representation spaces.
The total training objective is the sum of the standard Barlow Twins loss on the student's current projections and the distillation loss $\mathcal{L}_{\mathrm{distill}}$.

\newpage
\section{Extended chunking evaluation}
\label{appx:more_chunking}
In the main text Section~\ref{sec:coob_in_cl}, our empirical evaluation utilizes a $4$-chunk data-incremental setup. To demonstrate that the co-observation effect is not an artifact of this specific sequence length, but rather a robust property of the fragmented optimization, we extend our evaluation to longer sequences. We replicate the supervised learning setup on ImageNet-100, but partition the global training dataset uniformly into $8$ and $12$ identically distributed chunks, respectively. The fixed iteration budget is scaled proportionally to ensure the total optimization steps across the entire sequence remain consistent with the $4$-chunk and joint-training baselines.
\begin{figure}[!ht]
    \centering
    \begin{subfigure}{\textwidth}
        \centering
        \hspace*{5em}
        \includegraphics[width=0.8\linewidth]{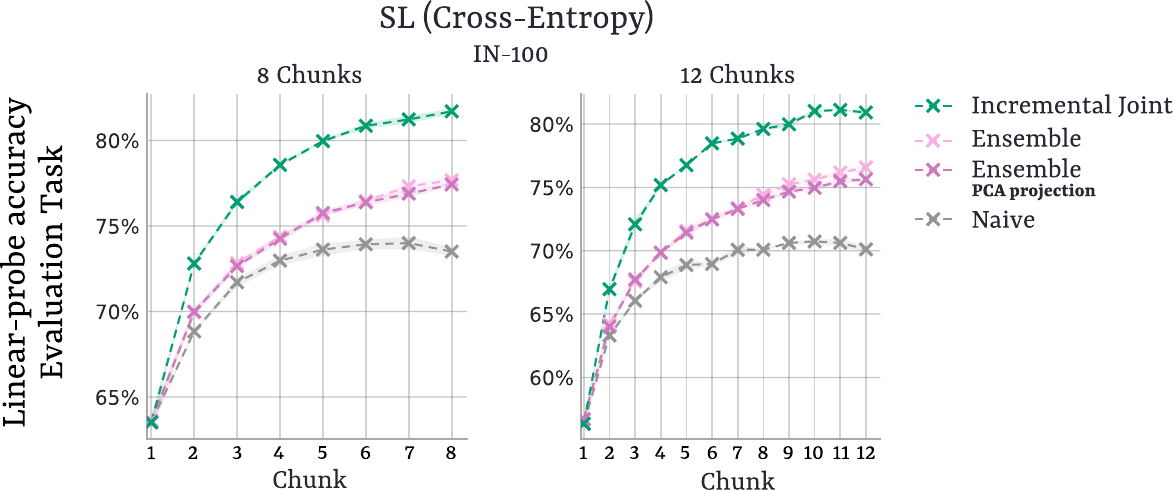}
        %\caption{}
    \end{subfigure}
    \ \\[1.5em]
    \begin{subfigure}{\textwidth}
        \centering
        \hspace*{5em}
        \includegraphics[width=0.8\linewidth]{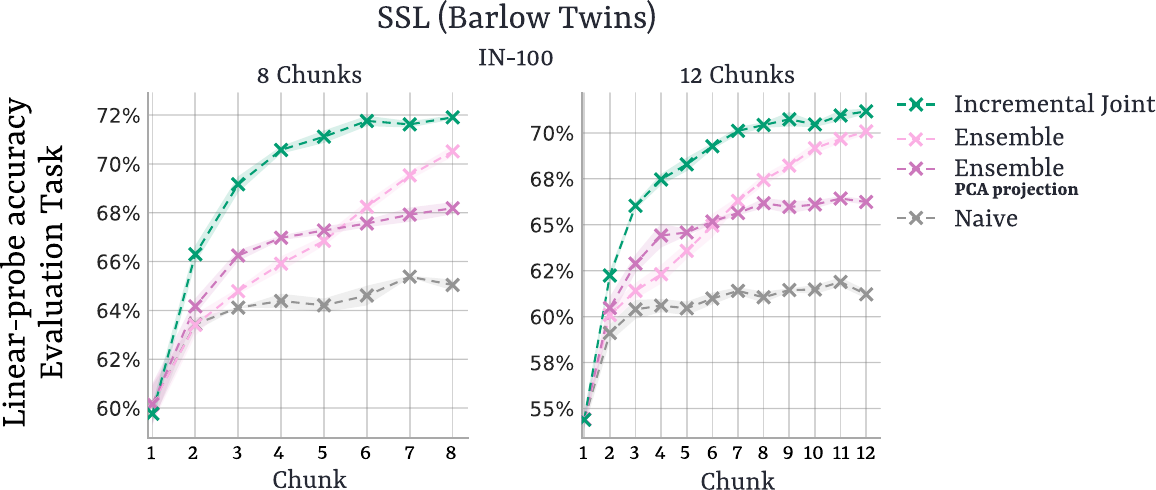}
        %\caption{}
    \end{subfigure}
    \caption{The co-observation effect across extended sequences. Linear probing accuracy, reported as the mean and standard error over 3 seeds, for supervised learning on ImageNet-100 partitioned into $8$ chunks (left) and $12$ chunks (right). The performance gap between the diagnostic ensemble (perfect retention) and incremental joint (simultaneous observation) persists consistently across longer sequences.}
    \label{fig:SL_ch8_and_12}
\end{figure}

As illustrated in Figure~\ref{fig:SL_ch8_and_12}, increasing the granularity of the data partitions does not alleviate the representational deficit. Across both the 8-chunk and 12-chunk sequences, the perfect-retention ensemble baseline successfully mitigates catastrophic forgetting relative to the naive sequential model, yet consistently falls short of incremental joint training.
This sustained gap confirms that the co-observation effect % -- the necessity of simultaneously observing data to synthesize optimal cross-distribution features -- 
remains a fundamental limiting factor regardless of the specific length or fragmentation of the continual learning sequence.

\newpage
\section{Numerical results}
\label{app:numerical_results}

\vspace*{3cm}
\begin{table*}[htbp]
  \captionsetup{justification=centering}
  \centering
  \small
  \caption{SL-CIFAR100 - LP-accuracy[\textbf{\%}] CIFAR100.\\ Runs reported as mean $\pm$ standard error over $3$ runs.}
  \vspace{0em}
  \begin{tabular}{lcccc}
    \toprule
    Method & Chunk 1 & Chunk 2 & Chunk 3 & Chunk 4 \\
    \midrule
    Naive            & $60.80{\scriptstyle \pm 0.31}$ & $63.54{\scriptstyle \pm 0.17}$ & $63.92{\scriptstyle \pm 0.20}$ & $64.64{\scriptstyle \pm 0.16}$ \\
    Ensemble         & $60.78{\scriptstyle \pm 0.30}$ & $65.93{\scriptstyle \pm 0.22}$ & $68.19{\scriptstyle \pm 0.37}$ & $70.39{\scriptstyle \pm 0.04}$ \\
    Ensemble PCA     & $60.83{\scriptstyle \pm 0.37}$ & $65.81{\scriptstyle \pm 0.12}$ & $68.12{\scriptstyle \pm 0.31}$ & $70.25{\scriptstyle \pm 0.06}$ \\
    Incremental Joint & $61.15{\scriptstyle \pm 0.16}$ & ${68.22}{\scriptstyle \pm 0.03}$ & ${72.45}{\scriptstyle \pm 0.03}$ & ${74.29}{\scriptstyle \pm 0.31}$ \\
    Joint (offline) & -- & -- & -- & $75.18{\scriptstyle \pm 0.55}$ \\
    \bottomrule
  \end{tabular}

  \vspace{2em}

  \caption{SL-IN-100 - LP-accuracy[\textbf{\%}] IN-100\\
  Runs reported as mean $\pm$ standard error over $3$ runs.
  }
  \vspace{0em}
  \begin{tabular}{lcccc}
    \toprule
    Method & Chunk 1 & Chunk 2 & Chunk 3 & Chunk 4 \\
    \midrule
    Naive            & $73.24{\scriptstyle \pm 0.48}$ & $76.66{\scriptstyle \pm 0.16}$ & $77.98{\scriptstyle \pm 0.29}$ & $78.27{\scriptstyle \pm 0.21}$ \\
    Ensemble         & $73.22{\scriptstyle \pm 0.42}$ & $77.48{\scriptstyle \pm 0.16}$ & $79.20{\scriptstyle \pm 0.24}$ & $80.55{\scriptstyle \pm 0.10}$ \\
    Ensemble PCA     & $73.25{\scriptstyle \pm 0.43}$ & $77.47{\scriptstyle \pm 0.21}$ & $79.15{\scriptstyle \pm 0.20}$ & $80.57{\scriptstyle \pm 0.11}$ \\
    Incremental Joint & $73.24{\scriptstyle \pm 0.48}$ & ${79.58}{\scriptstyle \pm 0.12}$ & ${81.85}{\scriptstyle \pm 0.10}$ & ${82.75}{\scriptstyle \pm 0.15}$ \\
    Joint (offline) & -- & -- & -- & $83.68{\scriptstyle \pm 0.11}$ \\
    \midrule
    Replay ($m_t\!=\!1250$)  & $73.07{\scriptstyle \pm 0.22}$ & $76.99{\scriptstyle \pm 0.14}$ & $78.35{\scriptstyle \pm 0.63}$ & $79.41{\scriptstyle \pm 0.23}$ \\
    Replay ($m_t\!=\!6250$)  & $73.06{\scriptstyle \pm 0.22}$ & $77.67{\scriptstyle \pm 0.12}$ & $79.54{\scriptstyle \pm 0.13}$ & $80.55{\scriptstyle \pm 0.23}$ \\
    Replay ($m_t\!=\!12500$) & $73.25{\scriptstyle \pm 0.07}$ & $78.05{\scriptstyle \pm 0.07}$ & $80.68{\scriptstyle \pm 0.15}$ & $81.27{\scriptstyle \pm 0.24}$ \\
    \midrule
    LwF              & ${73.35}{\scriptstyle \pm 0.23}$ & $77.41{\scriptstyle \pm 0.15}$ & $79.44{\scriptstyle \pm 0.22}$ & $79.93{\scriptstyle \pm 0.25}$ \\
    \bottomrule
  \end{tabular}

  \vspace{2em}
\end{table*}

\begin{table*}[htbp]
  \captionsetup{justification=centering}
  \centering
  \small

  \caption{Barlow-IN-100 - LP-accuracy[\textbf{\%}] IN-100\\
  Runs reported as mean $\pm$ standard error over $3$ runs.
  }
  \vspace{0em}
  \begin{tabular}{lcccc}
    \toprule
    Method & Chunk 1 & Chunk 2 & Chunk 3 & Chunk 4 \\
    \midrule
    Naive            & $67.47{\scriptstyle \pm 0.38}$ & $69.04{\scriptstyle \pm 0.46}$ & $69.64{\scriptstyle \pm 0.14}$ & $69.25{\scriptstyle \pm 0.06}$ \\
    Ensemble         & $67.56{\scriptstyle \pm 0.29}$ & $69.81{\scriptstyle \pm 0.29}$ & $70.37{\scriptstyle \pm 0.43}$ & $71.25{\scriptstyle \pm 0.19}$ \\
    Ensemble PCA     & $67.57{\scriptstyle \pm 0.37}$ & $70.20{\scriptstyle \pm 0.30}$ & $71.43{\scriptstyle \pm 0.30}$ & $71.53{\scriptstyle \pm 0.31}$ \\
    Incremental Joint & $67.19{\scriptstyle \pm 0.45}$ & ${71.61}{\scriptstyle \pm 0.36}$ & ${73.28}{\scriptstyle \pm 0.34}$ & ${73.50}{\scriptstyle \pm 0.41}$ \\
    Joint (offline) & -- & -- & -- & $76.42{\scriptstyle \pm 0.40}$ \\
    \bottomrule
    \midrule
    Replay ($m_t\!=\!1250$)  & $67.63{\scriptstyle \pm 0.42}$ & $69.25{\scriptstyle \pm 0.24}$ & $69.83{\scriptstyle \pm 0.14}$ & $69.68{\scriptstyle \pm 0.16}$ \\
    Replay ($m_t\!=\!6250$)  & $67.47{\scriptstyle \pm 0.29}$ & $69.69{\scriptstyle \pm 0.12}$ & $70.93{\scriptstyle \pm 0.17}$ & $71.63{\scriptstyle \pm 0.15}$ \\
    Replay ($m_t\!=\!12500$) & ${67.78}{\scriptstyle \pm 0.50}$ & $70.93{\scriptstyle \pm 0.04}$ & $71.99{\scriptstyle \pm 0.28}$ & $72.59{\scriptstyle \pm 0.09}$ \\
    \midrule
    CaSSLe           & $67.62{\scriptstyle \pm 0.25}$ & $69.96{\scriptstyle \pm 0.39}$ & $71.66{\scriptstyle \pm 0.28}$ & $72.07{\scriptstyle \pm 0.23}$ \\
    \bottomrule
  \end{tabular}

  \vspace{2em}

  \caption{IJEPA-IN-100 - LP-accuracy[\textbf{\%}] IN-100\\
  Runs reported as mean $\pm$ standard error over $6$ runs for Naive, Ensemble (PCA),\\ and (Incremental) Joint. And over $3$ runs for Replay.
  }
  \vspace{0em}
  \begin{tabular}{lcccc}
    \toprule
    Method & Chunk 1 & Chunk 2 & Chunk 3 & Chunk 4 \\
    \midrule
    Naive            & $55.87{\scriptstyle \pm 0.27}$ & $55.00{\scriptstyle \pm 0.38}$ & $54.63{\scriptstyle \pm 0.63}$ & $53.83{\scriptstyle \pm 0.60}$ \\
    Ensemble         & $55.88{\scriptstyle \pm 0.26}$ & $58.33{\scriptstyle \pm 0.41}$ & $59.11{\scriptstyle \pm 0.51}$ & $59.82{\scriptstyle \pm 0.46}$ \\
    Ensemble PCA     & $55.91{\scriptstyle \pm 0.26}$ & $57.08{\scriptstyle \pm 0.55}$ & $58.12{\scriptstyle \pm 0.67}$ & $58.83{\scriptstyle \pm 0.69}$ \\
    Incremental Joint & $55.80{\scriptstyle \pm 0.28}$ & $59.59{\scriptstyle \pm 0.32}$ & $61.83{\scriptstyle \pm 0.62}$ & $62.64{\scriptstyle \pm 0.55}$ \\
    Joint (offline) & -- & -- & -- & $69.35{\scriptstyle \pm 0.38}$ \\
    \midrule
    Replay ($m_t\!=\!1250$)  & $55.93{\scriptstyle \pm 0.69}$ & $56.18{\scriptstyle \pm 0.34}$ & $56.28{\scriptstyle \pm 0.88}$ & $54.79{\scriptstyle \pm 0.99}$ \\
    Replay ($m_t\!=\!6250$)  & $55.85{\scriptstyle \pm 0.31}$ & $57.07{\scriptstyle \pm 0.71}$ & $57.29{\scriptstyle \pm 0.37}$ & $56.59{\scriptstyle \pm 0.65}$ \\
    Replay ($m_t\!=\!12500$) & ${56.56}{\scriptstyle \pm 0.10}$ & $58.72{\scriptstyle \pm 0.06}$ & $59.41{\scriptstyle \pm 0.67}$ & $60.21{\scriptstyle \pm 0.99}$ \\
    \bottomrule
  \end{tabular}
\end{table*}

\end{document}